\documentclass[12pt]{article}

\usepackage{sbc-template}
\usepackage{graphicx,url}
\usepackage[utf8]{inputenc}

\usepackage{cite}
\usepackage{graphicx}
\usepackage{pgfplots}
\usepackage{multirow}
\usepackage{amsmath}
\usepackage{hyperref}

\pgfplotsset{scaled y ticks=false}
     
\title{PLSUM: Generating PT-BR Wikipedia by Summarizing Multiple Websites}

\author{André Seidel Oliveira\inst{1, [0000-0001-6551-6911]}, 
        Anna H. Reali Costa\inst{1, [0000-0001-7309-4528]}}

\address{Data Science Center (c$^2$d) - Universidade de São Paulo (USP)\\ São Paulo (SP), Brazil
\email{\{andre.seidel, anna.reali\}@usp.br}\\
\url{http://c2d.poli.usp.br/}}

\begin{document} 

\maketitle

\begin{abstract}
Wikipedia is an important free source of intelligible knowledge.
Despite that, Brazilian Portuguese Wikipedia still lacks descriptions for many subjects.
In an effort to expand the Brazilian Wikipedia, we contribute PLSum, a framework for generating wiki-like abstractive summaries from multiple descriptive websites.
The framework has an extractive stage followed by an abstractive one.
In particular, for the abstractive stage, we fine-tune and compare two recent variations of the Transformer neural network, PTT5, and Longformer.
To fine-tune and evaluate the model, we created a dataset with thousands of examples, linking reference websites to Wikipedia.
Our results show that it is possible to generate meaningful abstractive summaries from Brazilian Portuguese web content.
\end{abstract}

\section{Introduction}
Wikipedia is a public domain encyclopedia project, being a vast source of intelligible information on many subjects.
It has more than 51 million articles in over 300 languages, and around 1 million of them written in Portuguese. 
Despite the great voluntary effort to maintain it, there are still many topics not covered, especially in languages other than English.
For example, if we assume that Wikipedia articles describe language-independent topics, a quick comparison between the current number of articles in Portuguese and English shows that Portuguese Wiki lacks at least 5 million topics\footnote{Wikipedia numbers extracted from \url{https://pt.wikipedia.org/wiki/Wikipedia}, visited in 15/08/2021.}.

In contrast, there are plenty of textual descriptions online for just about anything.
In this sense, automatic text summarization techniques could be applied to these descriptions, as a way of generating Wikipedia articles.
So, as a first step towards the goal of expanding Wikipedia into Brazilian Portuguese, we tackle the task of automatically generating the first section of a Wikipedia article, i.e., a \emph{lead}. 
The lead section of a Wikipedia article is the section before the table of contents and serves as a summary of the article.

In particular, we handle the task of automatically generating the lead section of Wikipedia articles from \emph{a set of descriptions from the web}, so it is a \emph{multi-document abstractive summarization} (MDAS) task of reference websites.
It is called \emph{multi-document} because multiple documents are used to generate a single summary.
The name \emph{abstractive} comes from the fact that the output is generated by building new sentences, either by rephrasing or using new words from the model's vocabulary, rather than simply extracting important sentences from the documents, as is done in extractive summarization~\cite{gupta2019abstractive}.

An inspiring work for our investigation is WikiSum \cite{liu2018generating}, as it also addresses the generation of the lead section of Wikipedia through the summarization of websites.
We deepen their investigation of Transformer neural networks \cite{vaswani2017attention} in the MDAS domain by applying a \emph{pre-trained} model.
Transformer networks \cite{vaswani2017attention} is a model inspired by the success of attention in encoder-decoder neural networks. It is entirely based on attention heads that process input vector blocks in parallel.
The concept behind pre-training is to train a model on self-supervised language modeling tasks so that it learns contextual representations of sentences that can be further specialized with fine-tuning on specific tasks.
Indeed, pre-trained Transformers are the state-of-the-art for numerous seq2seq tasks \cite{raffel2019exploring}.

So, we present PLSum, the Portuguese Long Summarizer, an MDAS framework specialized in the generation of wiki-like Portuguese descriptions for multiple documents, based on Transformer networks.
Our contribution in this paper is twofold:
(1) To the best of our knowledge, this is the first work to apply Transformer networks for MDAS in Brazilian Portuguese.
In particular, we fine-tune and compare recently created Transformer-based models, notably PTT5 \cite{ptt5_2020} that is pre-trained on Portuguese data.
(2) We also release the \emph{BRWac2Wiki} dataset, automatically generated from thousands of pairs $\langle websites, Wikipedia \rangle$, which is a milestone for the Portuguese MDAS.
Results of our abstractive summarization framework show that it is possible to generate \emph{comprehensive} and \emph{long} summaries from reference websites.

The remainder of this paper is organized as follows: 
Section \ref{sec:related_w} describes works related to multi-document summarization for both English and Portuguese. 
Section \ref{sec:methods} presents our proposal, PLSum.
Then, in Section \ref{sec:pt_wikisum}, BRWac2Wiki dataset characteristics are presented.
Section \ref{sec:experiments} describe our experiments.
Finally, the results and conclusion are shown in Sections \ref{sec:results} and \ref{sec:conclusion}, respectively.

\section{Related Work}
\label{sec:related_w}
Despite the success of neural networks applied to seq2seq tasks (i.e. that receives a sequence of symbols as input, such as words or punctuation, and outputs another sequence) \cite{rush2015neural, raffel2019exploring}, until 2017 multi-document summarization was mostly explored by sentence extraction techniques.
Arguably this happened because of the lack of datasets with enough examples for neural training without over-fitting \cite{lebanoff2018adapting}. 

Later, PG-MMR \cite{lebanoff2018adapting} circumvented the lack of training examples for MDAS by adapting a model trained for headline generation on single-document summarization.
It applied an encoder-decoder Long Short-term Memory (LSTM) \cite{hochreiter1997long} Recurrent Neural Network (RNN) with attention \cite{bahdanau2014neural} and Pointer-generator (PG) \cite{see2017get}, a technique that enables the neural model to copy words from the input. 
The encoder-decoder RNN is a two-step architecture that first encodes input sequences in a continuous vector space, to later decode these vectors into new tokens, step by step until an output sequence is formed. 
Attention vectors are stacked on each stored output of the encoder to correlate them with each decoding step.
While they managed to perform MDAS, the fact that this model do not consider the input is from multiple-documents is a drawback, as it do not manage redundant information.

At the same time, WikiSum \cite{liu2018generating} was created: a model that generates wiki-like summaries in English from multiple documents.
They also released an homonym dataset for training and validation, relating websites extracted from google searches to Wikipedia. 
By applying automatic crawling for such websites and wikis, they managed to generate a dataset with more than 1 million examples.
They applied a decoder-only Transformer with local attention that enabled long input sequences for the abstractive stage. 
We consider this work a milestone for MDAS and our main inspiration, but they do not apply pre-trained Transformers, that are the state-of-the-art for several natural language generation tasks.

As far as we know, there are no MDS systems in Brazilian Portuguese that use abstractive techniques, nor Transformer networks.
It is worth mentioning CSTSumm \cite{jorge2010sumarizaccao}, a framework that utilizes \textit{cross-document structure theory} (CST) annotations to perform extractive summarization in Portuguese from multiple assessment template-based operators.
CST is a type of semantic relationship mapping between documents about the same subject.
Later, RSumm \cite{ribaldo2013investigaccao} also investigated multi-document extractive summarization in Portuguese, with the use of graphs for relationship mapping between sentences and path-finding for summarization.
Table \ref{tab:models_capabilities} sum up PLSum features in comparison to the publications cited here.

\begin{table}[]
\small
    \centering
    \caption{Comparison of capabilities and domains for different MDAS research. PLSum is in pair with recent MDAS publications for English language.}
    \label{tab:models_capabilities}
    \begin{tabular}{l c c c c c}
        \hline
        & PG-MMR & WikiSum & CSTSumm & RSumm & \textbf{PLSum (ours)} \\
        \hline
        \textbf{Abstractive} & X & X &  & & X \\
        \textbf{Use Transformers} & & X &  & & X \\
        \textbf{Use pre-train} & & &  & & X \\
        \textbf{For Portuguese} & & & X & X & X \\
        \hline
    \end{tabular}
\end{table}



\section{PLSum: Portuguese Long Summarizer}
\label{sec:methods}
\begin{figure}
    \centering
    \includegraphics[width=0.9\textwidth]{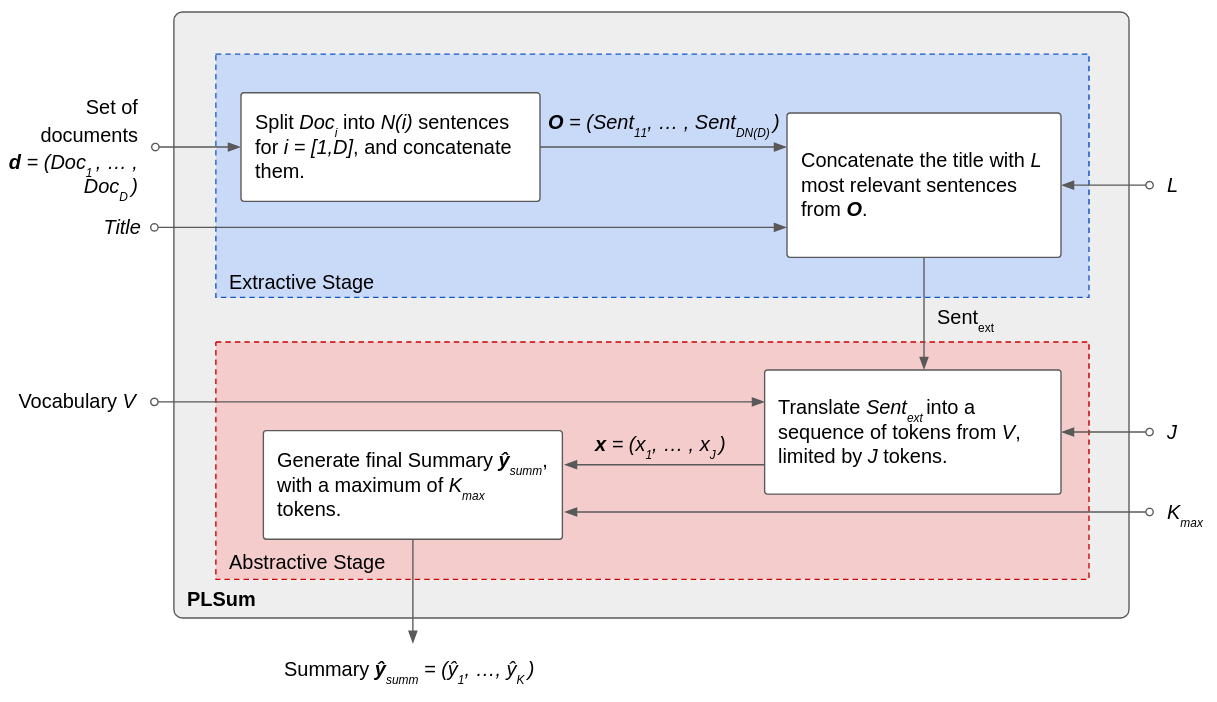}
    \caption{PLSum flowchart. Given the set of inputs (left side parameters); and the set of hyperparameters (right side parameters), PLSum filters relevant sentences and generates an authorial summary, ${\bf \hat{y}}_{summ}$.}
    \label{fig:PLSum}
\end{figure}

In this paper, we contribute with PLSum, a two-step architecture that combines extractive and abstractive stages.
The framework has as inputs (1) the desired title for the summary, $Title$, (2) a vocabulary of tokens, i.e numbers, punctuation, words, sub-words, $V = \{t_1, ..., t_{M}\}$, and (3) a set of documents related to this title, $\textbf{d} = \{Doc_1, ..., Doc_D\}$, where $M$ is the size of the vocabulary, and $D$ the number of documents.
PLSum reads the set of documents $\textbf{d}$ and returns a \emph{wiki-like summary} about the title. 
In particular, a summary is a sequence of tokens arbitrarily chosen from the vocabulary $V$ (i.e. abstractive summarization).
Figure \ref{fig:PLSum} illustrate the summarization process.

The extractive stage starts by dividing documents into sentences ending with a period, $Doc_i = (Sent_{i1}, ... , Sent_{iN(i)})$, where $Sent_i$ is a sentence with about 100 words and $N(i)$ is the number of sentences in $Doc_i$, which may vary from document to document.
The sentences from all documents are then grouped together into a super document, $\textbf{O} = (Sent_{11}, ..., Sent_{1N(1)}, ..., Sent_{D1}, ... , Sent_{DN(D)})$.
Then, it selects $L$ relevant sentences from the input set of documents $\textbf{O}$.
We apply TF-IDF \cite{ramos2003using} for the calculation of sentence relevance and selection.
\textbf{TF-IDF} stands for Term Frequency - Inverse Document Frequency and is a method for scoring terms relevance in a sentence $Sent_{ij}$ from a sequence of sentences $\textbf{O}$.
The score is defined as
\begin{equation}
    TFIDF_{term}(term, Sent_{ij}) = N_{term, Sent_{ij}} \  log( \frac{|\textbf{O}|}{N_{sentences-term}} ),
\end{equation}
where $N_{term, Sent_{ij}}$ is the number of times the term appears in $Sent_{ij} \in \textbf{O}$, and $N_{sentences-term}$ is the number of sentences of $\textbf{O}$ citing the term at least once.
TF-IDF assigns higher scores to terms that are frequent in $Sent_{ij}$, but not in the other sentences of $\textbf{O}$. On the other hand, it assigns lower scores to terms that are either frequent or uncommon in all sentences in $\textbf{O}$.
Sentence relevance is defined as the sum of $TFIDF_{term}(term, Sent_{ij})$ for every term in the title: 
\begin{equation}
    TFIDF_{sentence}(Sent_{ij}) = \sum_{term}^{title\ terms} TFIDF_{term}(term, Sent_{ij}).
\end{equation}

The $L$ most relevant sentences are then selected and concatenated with the title $Title$ in a single output sentence, $Sent_{ext}$, following descending order of relevance and with a separator token $[SEP]$ between them:
\begin{equation} \label{eq:out-ext}
Sent_{ext} = Title \ \text{{\footnotesize [SEP]}} \ \textit{Sent$_1$} \ \text{{\footnotesize [SEP]}} \ ... \textit{Sent$_i$} \ ... \ \text{{\footnotesize [SEP]}} \textit{Sent$_L$}.
\end{equation}

The abstractive stage starts by transforming numbers, punctuation, words, sub-words, from $Sent_{ext}$ into a sequence of known tokens from $V$, $\textbf{x} = (x_1, ..., x_J)$.
The input $\textbf{x}$ is limited to $J$ tokens, and this is a hyperparameter.
Then, the abstractive summarization objective is to generate another sequence of tokens from $V$,
${\bf \hat{y}}_{summ} = (\hat{y}_1, ..., \hat{y}_{K})$.
The number of tokens in the output sequence, $K$, is defined by the abstractive model (i.e. it may vary) and is limited by $K_{max}$, another hyperparameter.

In this paper, we compare two Transformer architectures for the abstractive stage: PTT5 \cite{ptt5_2020}, a checkpoint of T5 \cite{raffel2019exploring} pre-trained on a Brazilian Portuguese corpus; and Longformer \cite{beltagy2020longformer} encoder-decoder, an implementation of Transformer capable of processing longer documents, but without pre-training (see next section for an explanation of models).

\noindent \textbf{PTT5} \cite{ptt5_2020} is a seq2seq Transformer, based on T5 \cite{raffel2019exploring}, pre-trained on the Brazilian Portuguese corpus BrWac \cite{brwac}.
The model architecture is similar to the original encoder-decoder Transformer \cite{vaswani2017attention}, in which several blocks of self-attention layers and feedforward networks are concatenated. 
Attention output is defined as a function of key (Key), query (Query), and value (Value) vectors of dimension $dim$. 
In this case, the attention function is the ``scaled dot product", defined as:
\begin{equation}
    Attention(\text{Key, Query, Value}) = \frac{softmax(\text{Query}\cdot \text{Key}^T)}{\sqrt{dim}}. \text{Value}
\end{equation}
Attention layers might be (1) encoder-to-decoder attention, with Query and Value from the last encoder layer and Key from the last decoder layer, or (2) encoder-to-encoder self-attention, i.e Key, Query, and Value from the last encoder layer.
Also, the model uses parallel ``heads'' for computing attention, on each block the input Key, Query, and Value matrices are split and processed by independent attention ``heads''.
As the attention heads are order-independent, a relative position embedding is added to token embedding before processing.
PTT5 was pre-trained on BrWac for masked language modeling, where sentence tokens are replaced with a mask so that the model has to guess them.
We applied the ``base'' model\footnote{Available at \url{https://huggingface.co/unicamp-dl/ptt5-base-portuguese-vocab}.}, with 220M trainable parameters.
The base version has 12 layers across encoder and decoder, and 12 attention heads.

\noindent \textbf{Longformer encoder-decoder} \cite{beltagy2020longformer} is another variation of Transformer capable of processing longer inputs with the same model size than PTT5.
Among other minor changes, Longformer exchanges global self-attention blocks (attention correlation across all tokens) by local attention, where self-attention is computed in a sliding window around the central token. 
Also, dilated sliding window is applied, where the sliding window skips consecutive tokens to increase the receptive field without increasing computation. 
Despite the ability to have longer inputs, Longformer does not have a pre-trained checkpoint for Portuguese yet.
Since transfer learning on seq2seq tasks is reported to be effective \cite{raffel2019exploring}, the lack of pre-training is a drawback that might leverage the input size advantage.
We also applied the ``base'' version of Longformer encoder-decoder\footnote{Available at \url{https://huggingface.co/allenai/led-base-16384}.}, with similar characteristics to PTT5: it has 6 layers on the encoder, 6 on the decoder, 12 attention heads on both, and sliding window attention of 256 across every layer. 
Also, we follow the recommendation from the authors and apply global attention only to the first token.

\subsection{Alternative Extractive Techniques for Ablation Studies }
\label{sec:alt-ext}
We follow WikiSum \cite{liu2018generating}, and define two other extractive stages to perform ablation studies.

\begin{itemize}
    \item \textbf{Random:} is a technique that randomly chooses $L$ sentences from $\textbf{O}$. It is used to study how the abstractive system behaves without an adequate extractive stage. This technique serves to define a lower limit for the extractive stage.
    
    \item \textbf{Cheating:} is a technique whose sentence score is calculated by recalling the \textit{2-grams} between the sentences and the $target$ Wikipedia summary for the given title:
\end{itemize}

\begin{equation}
    Score_{Cheating} = \frac{\text{\textit{2-grams}}(Sent_i) \cap \text{\textit{2-grams}}(target)}{\text{\textit{2-grams}}(target)}.
\end{equation}

The cheating score exclusively accounts for the target summary as a way to generate the best possible extractive stage.
It is supposed to answer the question of how much better the extractive method can be and serve as an upper limit.

\section{BRWac2Wiki: Multi-document Summarization Dataset}
\label{sec:pt_wikisum}
To train the supervised models of the abstractive phase, as well as to validate the complete framework, we created the BRWac2Wiki dataset \footnote{See \url{https://github.com/aseidelo/BrWac2Wiki} for instructions on how to get the data.}. 
BRWac2Wiki is a structured dataset relating Brazilian websites to Wikipedia summaries.
Each row has a \emph{Wikipedia title}, the according Wikipedia summary, and a list of documents related to the title.

The documents are website texts from BrWac corpus \cite{brwac} that \emph{cite every word from the title, in any order}.
BrWac is a freely available Brazilian website corpus, composed of a list of records containing \textit{url}, \textit{title}, \textit{text}, and \textit{docid} (an unique id for each website).
This criteria to relate documents and titles is vulnerable to ambiguous title words, but we consider it is up to the algorithm that will use the dataset to handle the content selection.
Also, we applied post-processing, excluding rows on the dataset that did not regard the following rules: 
(1) A maximum of 15 documents for each summary;
(2) A minimum of 1000 words in total on each set of documents;
(3) A minimum of 20 words in each Wikipedia summary.
Finally, (4) examples were also subjected to \textit{clone detection}, as defined in WikiSum:
\begin{equation}
    P_{clone} = \frac{\text{\textit{1-gram}}(Doc_i) \cap \text{\textit{1-gram}}(a)}{\text{\textit{1-gram}}(a)},
\end{equation}
where websites $Doc_i$ were compared to Wikipedia's summary $a$. Sentences with $P_{clone} > 0.5$ were excluded.


Table \ref{tab:ptwikisum_percentiles} details the dataset characteristics in percentiles.
One should interpret values associated with percentiles as the maximum values for the smaller $x\%$ of every feature. 
``Input size" field shows the size, in number of words, of the concatenated input websites, the ``Output size" is the analogous quantity for Wikipedia summary, and ``N. documents" is the number of websites per example.
As it can be seen, BRWac2Wiki output sizes range from around 30 up to 3846 words, while the input size range from around 8033 to over 1M words, also 80\% of the outputs have less than 168 words.

\begin{table}[htb]
    \centering
    \caption{Characteristics of BRWac2Wiki dataset in percentiles. Input and output sizes are in number of words.}
    \label{tab:ptwikisum_percentiles}
    \begin{tabular}{l c c c c c}
        \hline
        \textbf{Percentile (\%)} & 20 & 40 & 60 & 80 & 100 \\
        \hline
        \textbf{Input size} & 8033 & 24210 & 53424 & 114777 & 1268802 \\
        \textbf{Output size} & 30 & 49 & 86 & 168 & 3846 \\
        \textbf{N. documents} & 2 & 8 & 15 & 15 & 15 \\
        \hline
    \end{tabular}
\end{table}

Table \ref{tab:dataset_sizes} shows a comparison between BRWac2Wiki and other summarization datasets, for both English and Portuguese. 
For each dataset, column ``\# ex." shows the number of examples, ``\# docs/summ." shows the maximum number of input documents in an example, and ``Task" is a brief description of the challenge.
Although it has 20 times fewer examples than WikiSum \cite{liu2018generating}, BRWac2Wiki is 2000 times larger than the other Portuguese MDAS dataset, CSTNews \cite{leixo2008cstnews}, and is comparable in size to CNN/Daily Mail \cite{hermann2015teaching} single-document summarization dataset.

\begin{table}[]
    \centering
    \caption{Comparison between BRWac2Wiki and other summarization datasets. The column \# ex. shows the number of input/output examples and \# docs/summ. is the maximum number of documents per summary.}
    \label{tab:dataset_sizes}
    \begin{tabular}{l l c c}
        \hline
        & \textbf{Task} & \textbf{\# ex.} & \textbf{\# docs/summ.} \\
        \hline
        CNN/Daily Mail (EN) & Gen. news highlight & 312K & 1 \\
        WikiSum (EN) & Gen. EN wiki & 2M & over 1K\\
        CSTNews (PT) & Gen. news summaries & 50 & 3 \\
        \hline
        \textbf{BRWac2Wiki (PT) (ours)} & Gen. PT wiki & 114K & 15 \\
        \hline
    \end{tabular}
\end{table}

\section{Methodology}
\label{sec:experiments}

In this section we describe the metrics we use to compare experimental results, then we discuss and describe the hyperparameters used and finally we detail the experimental part.

\subsection{Metrics for Automatic Evaluation}
Following other summarization publications, we use the ROUGE \cite{lin2004rouge} metric for automatic evaluation and comparison of models.
ROUGE is based on \textit{n-grams matching} between predicted and target summaries, and bounded between 0 and 1.
ROUGE is calculated by comparing each $\langle prediction, target \rangle$ pair, and the results are displayed as precision (P), recall (R), and F1 measure (F) percentages.
In particular, we apply three common versions for summarization, ROUGE-1, ROUGE-2, and ROUGE-L:

\begin{itemize}
    \item \textbf{ROUGE-$\langle N \rangle$ (R$\langle N \rangle$)}: computes P/R/F of $\langle N \rangle$-grams on the predicted summary, i.e. ROUGE-$\langle N \rangle$ is the number of targets $\langle N \rangle$-grams that match the predicted summary over the total amount of $\langle N \rangle$-grams in the target.
    \item \textbf{ROUGE-L (RL)}: computes P/R/F over n-grams matches, where n is the longest common subsequence (LCS) between target and predicted summaries. LCS accounts for subsequence matches with any number or size of gaps between elements, provided that they follow the same order.
\end{itemize}

\subsection{Hyperparameters and Optimization}
BRWac2Wiki was divided into train, validation and test sets with 91722 (80\%), 11465 (10\%), and 11465 (10\%) respectively.

\begin{figure}
    \centering
    \begin{tikzpicture}[scale=0.90]
    
        \centering
        \begin{axis}[
            title={},
            xlabel={L},
            ylabel={R2 R (\%)},
            xmin=0, xmax=20,
            ymin=0, ymax=15,
            xtick={0, 2, 4, 6, 8, 10, 12, 14, 16, 18},
            ytick={0, 3.1, 6.9, 9.3, 11.1, 12.4, 13.5},
            legend pos=south east,
            ymajorgrids=true,
            grid style=dashed,
        ]
        \addplot[
            color=blue,
            mark=square,
            ]
            coordinates {
            (0,0)(2,3.1)(4, 6.9)(6, 9.3)(8,11.1)(10, 12.4)(12, 13.4)(14, 13.5)(16, 13.5)(18, 13.5)
            };
        \end{axis}
    \end{tikzpicture}
    \caption{ROUGE 2 recall of TF-IDF extractive stage on validation set for different values of $L$ (number of extracted sentences).}
    \label{plot:extractivexL}
\end{figure}
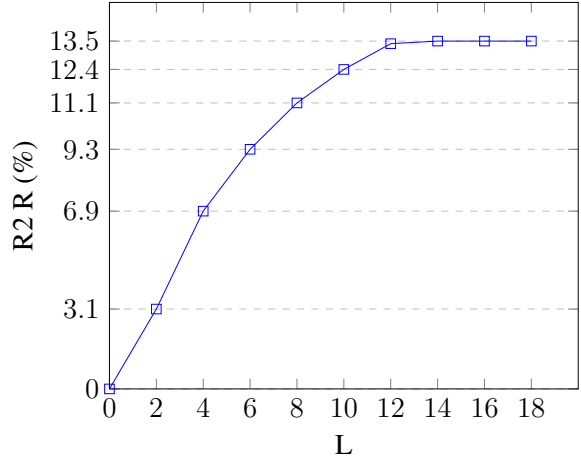

To define the desired input size for the abstractive models, we evaluated the recall of 2-grams (R2 R) between the extractive TF-IDF output and target on the validation set for different values of $L$ (number of selected sentences), from 0 to 20.
Figure \ref{plot:extractivexL} shows the results. 
One can notice that for $L > 12$ the R2 R measure does not increase over 13.5\%, which indicates that further increasing $L$ will not aggregate information for the abstractive model input.
Thus, we define our desired value for $L$ as 12, hence $J$ should be around 1200.
We trained the abstractive candidates on an Intel Xeon CPU @ 2.30GHz with Tesla V100 GPU.\footnote{Google Colab Pro specs at the time of publication.}
The limiting factor for input size was the GPU volume, so for PTT5 we were able to apply a maximum of $J = 768$, while for Longformer we managed to apply $J = 1024$.
In addition, we utilized the same output size $K_{max} = 256$ for both models, as 90\% of the target summaries on BrWac2Wiki dataset have less than 256 tokens, while the minimum size allowed was 20.

Both abstractive models (PTT5 and Longformer) were trained\footnote{Training scripts available at \url{https://github.com/aseidelo/wiki_generator}.} with similar hyperparameters:
Beam search with the number of beams equal 2, batch size of 16, and gradient accumulation steps equal 2.

\subsection{Experiments}
We performed 2 experiments to validate PLSum:

\noindent \textbf{Experiment 1:} is an ablation study of the TF-IDF extractive stage. 
We apply \emph{TF-IDF}, \emph{Random}, and \emph{Cheating} versions of the extractive stage on BRWac2Wiki test set, with L = 5.
Then, we compute ROUGE scores and compare results. The difference between \emph{TF-IDF} and \emph{Cheating} is an indication of how better the extractive stage could perform, while \emph{Random} is a random baseline.

\noindent \textbf{Experiment 2:} we fine-tune and compare four pipelines of PLSum, also with ROUGE automatic evaluation on BRWac2Wiki test set.
The compared pipelines are:
\begin{itemize}
    \item \textbf{TF-IDF + LF (Longformer), J = 1024:} the model with the bigger receptive field, but without pre-train and local attention;
    \item \textbf{TF-IDF + PTT5, J = 768:} the maximum receptive field possible for PTT5, pre-trained on Portuguese and with global attention;
    \item \textbf{TF-IDF + PTT5, J = 512:} is analogous to (2), but with smaller receptive field;
    \item \textbf{Random + PTT5, J = 768:} we apply Random as the extractive stage with PTT5 as the abstractive stage to asses the influence of the TF-IDF extractive in the complete framework.
\end{itemize}

Our results are reported in terms of the minimum, mean, and maximum f-measure (F) of ROUGE 1, 2, L, on test set samples. 
To estimate the boundary values (min., max.), we applied a bootstrap re-sampling on the test set, generating ROUGE scores for 1000 random samples.
Then, we considered the 2.5\% and 97.5\% percentiles of the ROUGE scores as the minimum and maximum boundaries respectively, i.e., a confidence interval of 95\%. 

\section{Results}
\label{sec:results}
The results of \textbf{Experiment 1} are shown in the rows associated to the ``Extractive" type, in Table  \ref{tab:pipelines_comparisson}.
The extractive methods have the expected performance for all ROUGE scores, that is $R_{Random} < R_{TF-IDF} < R_{Cheating}$.
\emph{TF-IDF} R1, R2, and RL F1 mean scores are 1.5, 1.1 and 0.9 above \emph{Random}, and 1.9, 2.6 and 0.8 bellow \emph{Cheating}, respectively.
Thus, while \emph{TF-IDF has some effectiveness as an extractive method} for Brazilian Portuguese sentences, it could still be improved if compared to the \emph{Cheating} method.

\begin{table}[]
    \small
    \centering
    \caption{ROUGE scores for the extractive stage and different abstractive pipelines of PLSum. Extractive models with $L = 5$ (number of 100 symbol sentences to extract) and Abstractive with $K_{max} = 256$ (maximum number of tokens on output). We consider a confidence interval of 95\% for boundary values within square brackets.}
    \label{tab:pipelines_comparisson}
    \begin{tabular}{c l c c c}
        \hline
        \textbf{Type} & \textbf{Model} & \textbf{R1 F (\%)} & \textbf{R2 F (\%)} & \textbf{RL F (\%)} \\
        \hline
        \multirow{3}{4em}{\textbf{Extractive}} & Random & 16.3 [16.1, 16.5] & 2.5 [2.5, 2.6] & 9.1 [9.1, 9.2]\\
        & TFIDF & 17.8 [17.7, 18.0] & 3.6 [3.5, 3.6] & 10.0 [9.9, 10.1] \\
        & Cheating & \textbf{19.7} [19.5, 19.9] & \textbf{6.2} [6.1, 6.2] & \textbf{10.8} [10.7, 10.8] \\
        \hline
        \multirow{4}{4em}{\textbf{Abstractive}} & TFIDF + LF, J=1024 & 19.7 [19.6, 19.8] & 9.2 [9.0, 9.5] & 19.8 [19.6, 20.0] \\
        & TFIDF + PTT5, J=768 & 32.0 [31.7, 32.4] & 14.9 [14.6, 15.2] & 25.6 [25.2, 25.9] \\
        & TFIDF + PTT5, J=512 & \textbf{33.4} [33.1, 33.6] & \textbf{16.4} [16.1, 16.6] & \textbf{27.1} [26.9, 27.4] \\
        & Random + PTT5, J=768 & 29.2 [29.0, 29.6] & 12.7 [12.5, 13.1] & 23.5 [23.1, 23.8] \\
        \hline
    \end{tabular}
\end{table}

For \textbf{Experiment 2}, the ``Abstractive" section of Table \ref{tab:pipelines_comparisson} compares the results for the different pipelines of the complete framework.
The method with \emph{TF-IDF + PTT5} and $J = 512$ has the best performance on every score. 
Also, every model with PTT5 achieved higher ROUGE scores than the one with Longformer.
\emph{Random + PTT5} has smaller ROUGE scores than \emph{TF-IDF + PTT5}, as expected. 
\emph{TF-IDF + PTT5} with $J = 512$ has higher mean ROUGE scores than the one with $J = 768$ (R1, R2, and RL are 33.4\%, 16.4\%, and 27.1\% against 32.0\%, 14.9\%, and 25.6\%, respectively).
Finally, if we consider the minimum and maximum values for each metric, we see that, for a 95\% confidence interval, results are unambiguous, as the scores do not overlap.

Therefore, we reached the following conclusions from the experiments: 
\begin{enumerate}
    \item The TF-IDF extractive stage is important, as TF-IDF only and TF-IDF + PTT5 had better results than \emph{Random} only and \emph{Random + PTT5}, with the same set of hyperparameters (10.0\% and 27.1\% against 9.1\% and 23.5\% RL F, respectively);
    \item The extractive stage alone is not enough, as the full abstractive framework had considerable better results than the best possible extractive method (\emph{Cheating}) (27.1\% against 10.8\% RL F);
    \item PTT5 had considerably better results than Longformer (27.1\% against 19.8\%), probably due to the pre-training in Brazilian Portuguese data;
    \item For PTT5, an exposure to a smaller receptive field ($J = 512$) displayed slightly better results (1.5\% increase on RL F). This might be because the version with the smaller receptive field is less exposed to ambiguous information in less relevant input sentences.
\end{enumerate}

\subsection{Qualitative Discussion}
\label{seq:qualitative}
To make a qualitative analysis of PLSum, we categorized strengths and issues found on the abstractive summaries generated with the test set.
Table \ref{tab:my_label} shows examples and highlights the errors presented here.

On the strengths of PLSum, we can highlight that the model showed the ability to merge concepts from the input sentences without having been explicitly coded to do so.
For instance, the summaries for ``Mario de Andrade" and ``Santos Dumont" display the dates of birth and death between parenthesis (like Wikipedia) by either compressing a sentence or merging information from multiple sentences.
We can also emphasize that, for most abstractive summaries, the information was displayed in the correct order. For example, biographical content most often began with the full name, followed by the date of birth and death, and then the important events and accomplishments about the person described. 
Finally, we highlight that PLSum can read texts written in multiple styles and write a wiki-like summary.
This feature is a great advantage compared to sentence extraction techniques.

However, PLSum has some problems. Although most summaries have syntactic and semantic consistency, many of them display incorrect information, particularly related to quantities. Dates, geographic positions, population, and area of countries were found to be incorrect. This issue might be related to biases generated by a relatively small number of training examples.

\begin{table}[htb]
    \small
    \centering
    \caption{Examples of automatically generated summaries (Predicted Summary) for some titles (Title). Wikipedia leads are displayed at Target Summary column. We highlight issues in yellow and annotate the type.}
    \label{tab:my_label}
    \begin{tabular}{|l | p{5.2cm} |p{5.2cm}|}
        \hline
        \textbf{Title} & \textbf{Predicted summary} & \textbf{Target summary} \\
        \hline
        \scriptsize{\textbf{Feudalismo}} & \scriptsize{o feudalismo foi um modo de organização social e política baseado nas relações servo contratos servis. tem suas origens na decadência do império romano. o sistema feudal se caracterizou pela concessão de feudos, quase sempre em forma de terras e trabalho, em troca de proteção política e militar. esse contrato feito era selado por um juramento de fidelidade.} & \scriptsize{feudalismo compreende o sistema político , econômico e social que predominou na europa ocidental entre o início da idade média até a afirmação dos estados modernos , tendo seu apogeu entre os séculos xi e xiii . o conceito teórico foi criado nos séculos xvii e xviii (...)} \\
        \hline
        \scriptsize{\textbf{Santos Dumont}} & \scriptsize{alberto santos dumont (\colorbox{yellow}{cabangu} \textbf{(WRONG)}, 20 de julho de 1873 são paulo, 23 de julho, 1932) foi um aviador brasileiro. foi o inventor do 14 bis, o primeiro avião a voar no brasil, e o primeiro piloto a voar na história do país.} & \scriptsize{alberto santos dumont (palmira , guarujá , ) foi um aeronauta , esportista e inventor brasileiro . santos dumont projetou , construiu e voou os primeiros balões dirigíveis com motor a gasolina . (...) }\\
        \hline
        \scriptsize{\textbf{Mário de Andrade}} & \scriptsize{mário raul de moraes andrade (são paulo, 9 de outubro de 1893 \colorbox{yellow}{rio de} \colorbox{yellow}{janeiro, 10 de agosto} \textbf{(WRONG)} de 1945) foi um poeta, ensaísta, crítico literário, \colorbox{yellow}{poeta e ensaísta} \textbf{(REPETITION)} brasileiro.} & \scriptsize{mário raul de morais andrade (são paulo , 9 de outubro de 1893 são paulo , 25 de fevereiro de 1945) foi um poeta , romancista , musicólogo , historiador de arte , crítico e fotógrafo brasileiro . (...)} \\
        \hline
    \end{tabular}
\end{table}

\section{Conclusion}
\label{sec:conclusion}
In this work, we innovated by applying Transformers networks to generate abstractive summaries from multiple documents in the Brazilian Portuguese language.
Despite displaying some unreliable information, the model showed great synthesis capacity and was able to create long, comprehensive summaries, clearly in the style of Wikipedia.
In this way, we contribute not only a first step towards the automatic generation of articles for uncovered wiki topics, but also the means for future research to tackle the task of MDAS in the Portuguese language.

The results support the conclusions of previous work that abstractive models can generate better-written summaries than extractive ones.
Furthermore, they have the advantage of being able to adapt to different styles when fine-tuned.
Another major conclusion of this work is that pre-training is very effective and was more determinant than having a bigger input into the abstractive stage.

Interesting themes for future research are addressing the factual inaccuracy of such abstractive models. A possible solution, already applied to the English language, is the use of mixed extractive and abstractive techniques, such as Pointer-generator, so that the model can copy crucial information from the input.
Also, increasing the training dataset can lessen the effect of biases towards commonplaces. Another future work is the application of pre-training for Brazilian Portuguese in models with a receptive field superior to T5, such as Longformer, which can lead to more complete summaries. Finally, better extractive techniques could be explored in conjunction with abstractive models.

\section*{Acknowledgments}
This research was supported by \textit{Ita\'{u} Unibanco S.A.}, with the scholarship program of \textit{Programa de Bolsas Ita\'{u}} (PBI), and partially financed by the Coordenação de Aperfeiçoamento de Pessoal de Nível Superior (CAPES), Finance Code 001, and CNPQ (grant 310085/2020-9), Brazil.
Any opinions, findings, and conclusions expressed in this manuscript are those of the authors and do not necessarily reflect the views, official policy or position of the Itaú-Unibanco, CAPES and CNPq.

\bibliographystyle{sbc}
\bibliography{sbc-template}

\end{document}